\documentclass{article}
\usepackage{spconf,amsmath,graphicx, amsthm}
\usepackage{booktabs}
\usepackage{dingbat}
\usepackage{comment}
\usepackage{multirow}
\usepackage{enumitem}
\usepackage{setspace}

\title{Maximizing data efficiency for Cross-Lingual TTS Adaptation by Self-Supervised Representation Mixing and Embedding Initialization}
\name{Wei-Ping Huang, Sung-Feng Huang, Hung-yi Lee}
\address{
 National Taiwan University\\
 Graduate Institute of Communication Engineering \\
 r11942102@ntu.edu.tw, f06942045@ntu.edu.tw, hungyilee@ntu.edu.tw
}
\copyrightnotice{979-8-3503-0689-7/23/\$31.00~\copyright2023 IEEE}
\begin{document}
%
\maketitle
\begin{abstract}

This paper presents an effective transfer learning framework for language adaptation in text-to-speech systems, with a focus on achieving language adaptation using minimal labeled and unlabeled data. While many works focus on reducing the usage of labeled data, very few consider minimizing the usage of unlabeled data. By utilizing self-supervised features in the pretraining stage, replacing the noisy portion of pseudo labels with these features during fine-tuning, and incorporating an embedding initialization trick, our method leverages more information from unlabeled data compared to conventional approaches. Experimental results show that our framework is able to synthesize intelligible speech in unseen languages with only 4 utterances of labeled data and 15 minutes of unlabeled data. Our methodology continues to surpass conventional techniques, even when a greater volume of data is accessible. These findings highlight the potential of our data-efficient language adaptation framework.
\end{abstract}
\begin{keywords}
speech synthesis, transfer learning, cross-lingual, low-resource language, self-supervised features
\end{keywords}
\section{Introduction}
\label{sec:intro}

Text-to-speech (TTS) has made significant advancements in recent years through the application of deep learning techniques\cite{kim2021conditional, ren2022portaspeech, wang2017tacotron, shen2018natural, ping2018deep, sotelo2017char2wav}, resulting in the development of TTS models capable of synthesizing natural, human-like speech. However, training a high-fidelity TTS system typically requires a substantial amount of high-quality recordings aligned with their corresponding transcriptions. This requirement poses a challenge, particularly in languages with limited resources. Consequently, increasing data efficiency has become a major issue when developing advanced TTS systems for low-resource languages.

A common approach is transfer learning~\cite{yang2020towards, li2019bytes}, which takes advantage of the ease of collecting data from rich-resource languages. By training with data from multiple languages in advance, the model benefits from cross-lingual information and improved data efficiency when adapting to a new language. 
Another approach to increase data efficiency is to exploit unlabeled data of target languages~\cite{ren2019almost, ni2022unsupervised}. However, collecting a large unlabeled corpus for an unseen language incurs additional costs, and these resources may not even exist for endangered languages. 
While significant achievements have been made in reducing the usage of labeled data, very few works focus on minimizing unlabeled data usage to the greatest extent possible, exploring the limitations of building a TTS system for an unseen language.

Our work aims to explore language adaptation with both minimal labeled and unlabeled data from the target language. Moreover, we assume that the input is language-dependent, which poses additional challenges for language adaptation with high data efficiency. We propose a simple but effective method based on both transfer learning and pseudo-labeling, a well-known method to deal with unlabeled data. 
The major contribution of this paper is the new idea about the effective use of pseudo labels, while simultaneously maintaining the model's performance even when faced with inaccuracies in these labels.
Previous work~\cite{xu2020lrspeech, ren2023bag} has shown that applying filtering-based methods to discard a portion of pseudo labels based on a noisiness criterion is effective.
We propose replacing the noisy portions with self-supervised features to maximize the data efficiency instead of discarding these transcriptions. 
This approach utilizes complete unlabeled data and exploits more information during transfer learning. 
Furthermore, by incorporating an embedding initialization trick proposed in \cite{huang2022fewshot}, we achieve highly data-efficient language adaptation while maintaining naturalness. 

Experiments show that the proposed method can adapt the TTS model to a new language under extremely low-resource settings with only 4 utterances of labeled data and 15 minutes of unlabeled data, and continues to surpass conventional techniques when a greater volume of data is accessible.

\section{Related Work}
\label{sec:related}

\begin{table*}[t]
  \caption{The data setting applied by previous works when learning a new language. Note that Lrspeech\cite{xu2020lrspeech}* utilizes additional low-quality data in both paired and unpaired data.}
  \centering
  \begin{tabular}{lcccccccccc}
    \toprule
    \textbf{Method} & \cite{chung2018semisupervised} & \cite{zhang2020unsupervised} & \cite{Kim_2022} & \cite{ren2019almost} & \cite{ni2022unsupervised} & \cite{xu2020lrspeech}* & \cite{yang2020towards} & \cite{chen2019end} & \cite{huang2022fewshot} & Our work \\
    \midrule
    \textbf{Paired data} & 24min & 24min & 10min & 20min & 0 & 1.34hr & 6min & 15min & 30sec & 30sec\\
    \textbf{Unpaired speech} & 40hr & 24hr & 24hr & 24hr & 960hr & 11.8hr & 0 & 0 & 0 & 15min\\
    \textbf{Unpaired text} & \, & \, & \, & \checkmark & \checkmark & \checkmark & \, & \, & \, & \,\\
    \textbf{Rich-Resource Lang.} & \, &\, & \,& \, & \, & \checkmark & \checkmark & \checkmark & \checkmark & \checkmark \\
    \bottomrule
  \end{tabular}
  \label{tab:categorize}
\end{table*}

{\setstretch{1.0}
Various approaches have been explored to relax the data requirements for training TTS systems. Semi-supervised learning leverages additional unpaired speech or text data that does not require a significant cost to collect. \cite{chung2018semisupervised, zhang2020unsupervised, Kim_2022} improve data efficiency by introducing an unsupervised pretraining stage. \cite{ren2019almost} applies back translation, which learns ASR and TTS iteratively and boosts the performance of two tasks gradually. More recently, self-supervised learning(SSL)\cite{9086055, LIU2022100616} has gained popularity as it allows for the exploitation of unlabeled data by learning implicit structure information. \cite{saeki2023virtuoso, saeki2023learning} leverages self-supervised learning with a large amount of unspoken text and untranscribed speech to build a highly transferable TTS system. Self-supervised learning has also given rise to unsupervised ASR\cite{NEURIPS2021_ea159dc9, liu2022endtoend}, enabling researchers to create pseudo labels without any paired data and achieve fully unsupervised TTS\cite{ni2022unsupervised}.

For transfer-learning-based approaches, \cite{yang2020towards} trains a TTS model with a large, multilingual speech corpus consisting of 50 languages and demonstrates that model can efficiently adapt to a new language using only 6 minutes of paired data. To reduce input space mismatch between different languages during transfer learning, many works leverage shared features as input. \cite{gutkin2017uniform, gutkin2017areal, demirsahinunified, maniati2021cross} use phonological features(PF) and International Phonetic Alphabet(IPA)\cite{international1999handbook}, and \cite{he2021multilingual, li2019bytes} are based on bytes. On the other hand, some works avoid such handcrafted shared features and keep the input remain language dependent. \cite{chen2019end} proposes a phoneme representation mapping network between two languages, while \cite{huang2022fewshot} attempts to construct a unified phoneme representation across multiple languages. Transferring from rich-resource languages can also be seamlessly integrated with back translation. \cite{xu2020lrspeech} additionally applies filtering based on an attention score to handle noisy pseudo labels, while \cite{ren2023bag} applies voice conversion to normalize noisy data and utilizes several tricks to improve the stability and performance of back translation. 

We summarize the data resources used by previous work when learning a new language in Table~\ref{tab:categorize}. \cite{chung2018semisupervised, zhang2020unsupervised, Kim_2022, ren2019almost, ni2022unsupervised} do not utilize data from rich-resource languages but instead require a lot of unpaired data on the target language for training. In terms of data usage minimization, \cite{huang2022fewshot} reduces it to the most extreme scenario by utilizing only four utterances of paired data without any unpaired data. However, they reported that the resulting utterances were not natural enough and had an unnatural accent. In contrast, our method exploits both paired and unpaired data to achieve maximum data efficiency and is capable of synthesizing more natural speech.}

\section{Method}
\label{sec:method}

\subsection{Overview}
\label{section:overview}
We present a data-efficient framework for cross-lingual adaptation in phoneme-based TTS based on transfer learning and pseudo-labeling. Our framework introduces several techniques to enhance both the pretraining and fine-tuning stages. 

Before introducing the proposed framework, it is important to clarify the data usage. When adapting to a new language, we assume that we have access to \textbf{(a)} Paired data from other rich-resource languages, denoted as $D_{source}=(T_{source}, S_{source})$, \textbf{(b)} Paired data from the target language $D_{target}=(T_{target}, S_{target})$ and \textbf{(c)} Unpaired speech data from the target language $S^u_{target}$. We assume that phoneme boundaries are available.

\begin{figure*}[htb]
\begin{minipage}[b]{1.0\linewidth}
  \centering
  \centerline{\includegraphics[width=0.85\linewidth]{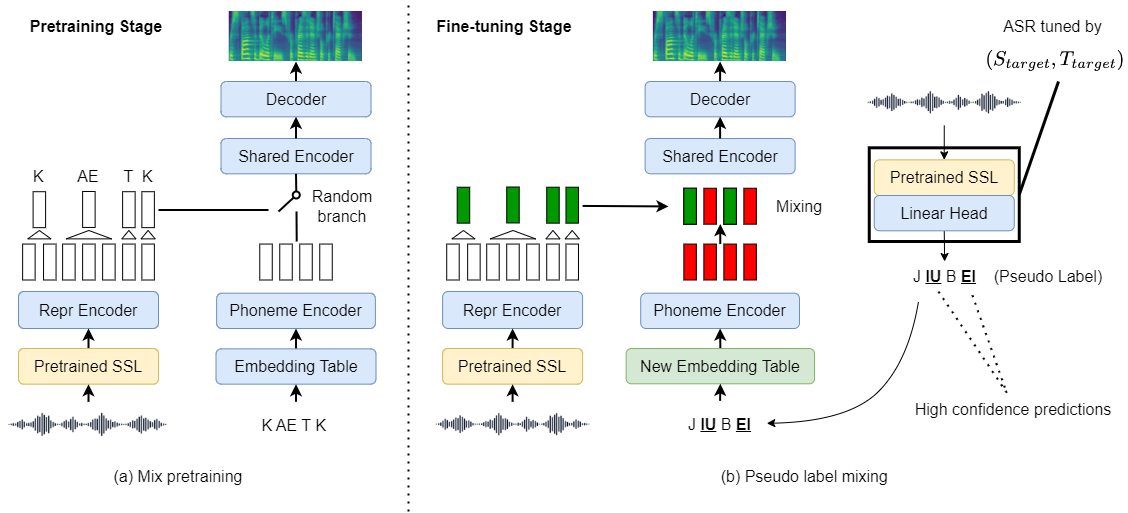}}
\end{minipage}
\caption{Illustration of the overall pipeline. (a) Mix pretraining. (b) Pseudo label mixing. We use $D_{source}$ for pretraining, and $D_{target}$ merged with pseudo corpus for fine-tuning.}
\label{fig:overview}
\end{figure*}

The overall pipeline is illustrated in Figure~\ref{fig:overview}. 
Section~\ref{section:MP} describes the pretraining stage illustrated in Figure~\ref{fig:overview}a. The model is trained using paired data from multiple rich-resource languages concurrently. We additionally input self-supervised features and jointly train the model to reconstruct the speech from these features with a random branch.
For the subsequent fine-tuning stage in Figure~\ref{fig:overview}b, we first generate pseudo transcripts $\widehat{T}^{u}_{target}$ for the unpaired speech $S^u_{target}$ with an SSL-based ASR system trained only on $D_{target}$ as in Section~\ref{section:PL}. Pseudo corpus $(\widehat{T}^u_{target}, S^u_{target})$ are then merged with $D_{target}$ to fine-tune the model's parameters. Section~\ref{section:PLM} introduces \textit{pseudo label mixing} to improve the fine-tuning stage. Finally, in Section~\ref{section:FSCL}, we further extend our method with an embedding initialization trick proposed by \cite{huang2022fewshot}.

\subsection{Mix Pretraining}
\label{section:MP}

In addition to synthesizing speech from the input, we train the model to reconstruct the speech from self-supervised features, where both tasks are optimized together. We utilize an additional representation encoder as shown in Figure~\ref{fig:overview}a. For the shared encoder, the input is randomly sourced from either the phoneme encoder or the representation encoder.

Frame-level self-supervised features are extracted with an off-the-shelf SSL model pretrained on other rich-resource languages. The SSL model is fixed through the training process. The representation encoder applies a weighted sum on the features from the different layers of the SSL model, followed by several transformer blocks. For every phoneme, we average the corresponding frame-level representations that belong to it according to provided phoneme boundaries, resulting in phoneme-level representations as in Figure~\ref{fig:overview}a.

Mix pretraining encourages the model to learn from both the distributions of self-supervised representations and supervised representations, which benefits the model by utilizing both types of representations during the pretraining stage.

\subsection{Pseudo Label Generation}
\label{section:PL}
The simplest way to leverage untranscribed speech in the fine-tuning stage is to generate pseudo transcripts with an additional ASR system. Previous studies, such as wav2vec2.0\cite{baevski2020wav2vec} and HuBERT\cite{hsu2021hubert}, have demonstrated that SSL models can be fine-tuned with an additional head to perform speech recognition or phoneme recognition, even with limited available data.

We fix the SSL model and fine-tune a linear head on top using only the paired data from the target language. To generate the phoneme sequence, we predict the most possible phoneme for each frame and then merge consecutive identical predictions into a single phoneme. 

\subsection{Pseudo Label Mixing}
\label{section:PLM}

Pseudo labels generated in Section \ref{section:PL} might contain many errors since our ASR is only fine-tuned on limited paired data. To deal with the presence of wrong labels, we proposed to mix outputs from the phoneme encoder and the representation encoder. We replace some portion of the phoneme encoder's output with the representation encoder's output based on the confidence score predicted by the ASR system.

Denote the phoneme encoder's output as $c_{phn}$ and the representation encoder's output as $c_{repr}$. Both of them are phoneme-level representations, and we index the $i$-th phoneme's representation as $c_{phn}^i$ and $c_{repr}^i$. Denote $s^i$ as the confidence score of the $i$-th phoneme predicted by the ASR system.  Since we merge consecutive identical predictions into a single phoneme, $s^i$ is calculated by averaging the confidences over those consecutive frames. Denote the mixed output as $c_{mix}$, our goal is to construct the mixing function $c_{mix}=f(c_{repr}, c_{phn})$. We propose two mixing functions.

\noindent {\bfseries Phoneme-level Mix.}\quad For every phoneme, this method decides whether to use $c_{phn}$ or $c_{repr}$ based on if its confidence score has surpassed a predefined threshold. The mixing function is
\begin{equation}
c_{mix}^i = f(c_{repr}^i, c_{phn}^i) = \begin{cases} 
c_{phn}^i,  & \text{if }s^i\geq \lambda \\
c_{repr}^i, & \text{otherwise},
\end{cases}
\end{equation}
where $\lambda$ is a hyperparameter.

\noindent {\bfseries Sentence-level Mix.}\quad For every sentence, this method decides whether to use $c_{phn}$ or $c_{repr}$ based on if its 
averaged confidence score has surpassed a predefined threshold. Denote the sequence length as $L$, the mixing function is
\begin{equation}
c_{mix} = f(c_{repr}, c_{phn}) = \begin{cases} 
c_{phn},  & \text{if }\frac{1}{L}\sum_i s^i\geq \lambda \\
c_{repr}, & \text{otherwise}.
\end{cases}
\end{equation}

\begin{figure}[htb]
\begin{minipage}[b]{1.0\linewidth}
  \centering
  \centerline{\includegraphics[width=1.0\linewidth]{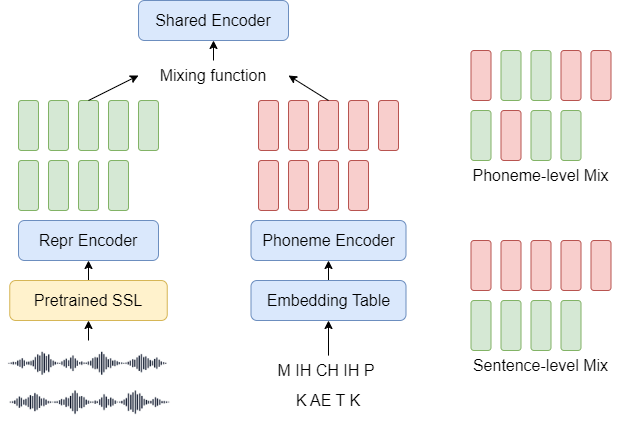}}
\end{minipage}
\caption{Illustration of proposed pseudo label mixing methods.}
\label{fig:pl-mixing}
\end{figure}

Figure~\ref{fig:pl-mixing} illustrates the difference between the two methods. Since the representation encoder is trained for reconstruction, its output will be perfectly aligned with the target speech. Our proposed mixing functions view low-confidence predictions as noisy parts and replace them with perfectly aligned features. Such replacement hence reduces the possible misalignment of input and output due to the wrong labels from the ASR system.

\subsection{Embedding Initialization Trick} 
\label{section:FSCL}

Each language has its own distinct phoneme set since we assume that input is language-dependent. Therefore when adapting to a new language, the phoneme embedding table is not transferable and needs to be reinitialized. \cite{huang2022fewshot} shows that under extremely low resource settings where only a few utterances of paired data are available, TTS adaptation might fail due to critical overfitting problems. They propose a trick that initializes the phoneme embedding table from a few utterances of paired data instead of randomly initializing it to deal with the problem.

We extend our method with the embedding initializing trick to further increase our data efficiency. The trick introduces a second pretraining stage that learns an \textit{embedding generator} to generate a phoneme embedding table from data. We take the representation encoder's output as the input of the embedding generator.

At the beginning of the fine-tuning stage, the embedding generator uses $D_{target}$ to generate a phoneme embedding table for initialization. We tune the embedding table, phoneme encoder, shared encoder, and decoder while fixing other modules when fine-tuning.

\section{Experiments}
\label{sec:format}

\subsection{Datasets}
We use 6 languages: English, Mandarin, French, German, Japanese, and Korean. Japanese and German are chosen for testing and the other 4 languages for training. We use LibriTTS\cite{zen2019libritts} train-clean-100 subset for English (\textit{en}), AISHELL-3\cite{shi2020aishell} for Mandarin (\textit{zh}), CSS10\cite{park2019css10} for French (\textit{fr}) and German (\textit{de}), KSS dataset for Korean (\textit{ko}), and JSUT\cite{sonobe2017jsut} for Japanese (\textit{jp}). We use Montreal Forced Aligner\cite{mcauliffe2017montreal} to generate phoneme boundaries, and every language is independently aligned with its own phoneme set. 10\% of the data are reserved as testing set.

\subsection{Training Setup}

Our TTS model is based on a modified version of multispeaker FastSpeech2\cite{ren2020fastspeech} from \cite{huang2022fewshot}. We use 2 transformer blocks for the representation encoder, phoneme encoder, and shared encoder. Hubert-large is chosen as the high-resource-pretrained SSL model in all experiments. Hubert-large is also used for pseudo label generation. Note that the simplicity of our proposed approach allows for easy generalization to other TTS architectures, not limited to FastSpeech2.

For the pretraining stage, we apply Adam optimizer\cite{kingma2014adam} with a learning rate of 0.001 for 50K iterations. Warmup for 2K iterations followed by an inverse square root learning rate decay is applied. We use a HifiGAN\cite{kong2020hifigan} vocoder checkpoint released by its authors to convert Mel-spectrograms back to audios for all experiments.

\subsection{Adaptation Strategy}
\label{section:strat}

\subsubsection{Combinations of Strategies}
We experiment on different combinations of adaptation strategies with techniques proposed in Section~\ref{section:PL} and Section~\ref{section:PLM}. For the pretraining stage, the adaptation strategy can apply \textit{mix pretraining} or not. For the fine-tuning stage, besides \textit{pseudo label mixing}, we introduce two baselines.

\noindent {\bfseries Sentence-level filter.}\quad Remove the sentence with averaged confidence lower than the threshold from the dataset. This is the common filter-based method.

\noindent {\bfseries Phoneme-level filter.}\quad Apply a mask on loss calculation. In FastSpeech2's case, Mel-spectrogram loss, pitch loss, energy loss, and duration loss are only calculated on frames corresponding to high-confidence phonemes.

Since \textit{pseudo label mixing} must be used together with \textit{mix pretraining}, combining methods from both stages results in six combinations in total. We abbreviate \textit{phoneme-level mix}, \textit{sentence-level mix}, \textit{phoneme-level filter}, and \textit{sentence-level filter} as \textbf{PM}, \textbf{SM}, \textbf{PF} and \textbf{SF}, respectively.

Instead of fixing a confidence threshold, we set the threshold according to a fixed \textit{psuedo label ratio}, which is defined as the ratio of predictions with confidence surpassing the threshold $\lambda$, to make different methods comparable. We experiment on 25\%, 50\%, 75\%, and 100\% pseudo label ratios. Note that under the 100\% case, $\lambda$ is set to 0 and all four tuning strategies become identical.

\subsubsection{Results}
\label{section:eval1}

We use character error rate (CER) and Mean of Opinion Score test (MOS) as our evaluation metrics. For the CER, we use Google Cloud API to perform speech recognition for synthesized recordings. CER is averaged over 5 runs to reduce the performance variance.

For each run, we randomly choose 16 utterances from the training set as labeled data while others were left as unlabeled data, which is about 5.7 hours for \textit{jp} and 6.7 hours for \textit{de}. We evaluate CER on 64 utterances chosen from the testing set. Following \cite{huang2022fewshot}, we ensure any phoneme that appears in the evaluation set also appears at least once in the labeled data. 

\begin{table}[t]
  \caption{CERs[\%] of different combinations of adaptation strategies. \textbf{MP} represents mix pretraining and \textbf{FT} represents fine-tuning. We abbreviate \textit{phoneme-level mix}, \textit{sentence-level mix}, \textit{phoneme-level filter}, and \textit{sentence-level filter} as \textbf{PM}, \textbf{SM}, \textbf{PF} and \textbf{SF}, respectively. "*" represents the best result in a single language.}
  \centering
  \begin{tabular}{ccc|cccc}
    \toprule
    \multirow{2}{*}{\textbf{Lang.}} & \multirow{2}{*}{\textbf{MP}} & \multirow{2}{*}{\textbf{FT}} & \multicolumn{4}{c}{\textbf{Pseudo Label Ratio}}\\
    & & & 25\% & 50\% & 75\% & 100\%\\
    \midrule
    \multirow{6}{*}{\textit{jp}} & \checkmark & PM & 21.66 & \textbf{17.47} & \textbf{17.23*} & \textbf{25.71}\\
    & \checkmark & PF & \textbf{21.25} & 19.79 & 19.17 & 25.71\\
    &  & PF & 21.27 & 19.21 & 19.52 & 28.64 \\
    
    & \checkmark & SM & 29.54 & 28.09 & 26.79 & 25.71\\
    & \checkmark & SF & 33.63 & 29.24 & 26.93 & 25.71\\
    &  & SF & 32.83 & 31.27 & 28.59 & 28.64 \\
    \midrule
    \multirow{6}{*}{\textit{de}} & \checkmark & PM & 15.23 & \textbf{11.31} & \textbf{9.44*} & 12.47\\
    & \checkmark & PF & \textbf{14.47} & 13.68 & 12.49 & 12.47\\
    &  & PF & 15.39 & 13.41 & 12.84 & \textbf{12.39} \\
    & \checkmark & SM & 13.88 & 12.1 & 12.05 & 12.47\\
    & \checkmark & SF & 16.77 & 14.06 & 12.21 & 12.47\\
    &  & SF & 17.97 & 14.37 & 13.4 & 12.39 \\
    \bottomrule
  \end{tabular}
  \label{tab:main-exp}
\end{table}

\begin{table}[t]
  \caption{MOS of different combinations of adaptation strategies. Pseudo label ratio is fixed at 75\%.}
  \centering
  \begin{tabular}{cc|cc}
    \toprule
    \textbf{MP} & \textbf{FT} & \textit{jp} & \textit{de}\\
    \midrule
    \checkmark & PM & \textbf{3.24} & \textbf{2.72}\\
    \checkmark & PF & 2.91 & 2.13\\
     & PF & 3.00 & 2.09 \\
    \checkmark & SM & 3.09 & 2.35\\
    \checkmark & SF & 2.87 & 2.27\\
     & SF & 2.84 & 2.02\\
    \midrule
    \multicolumn{2}{c|}{Ground Truth}  & 3.97 & 3.92\\
    \bottomrule
  \end{tabular}
  \label{tab:main-mos-exp}
\end{table}

Table~\ref{tab:main-exp} summarizes the CER results.
The combination of \textit{mix pretraining} and \textit{phoneme-level mix} achieves the best performance in both languages with a 75\% pseudo label ratio. We observe that phoneme-level-based methods are comparable to or better than their sentence-level-based counterparts. This suggests that controlling pseudo label usage at a fine-grained level is beneficial. Moreover, proposed \textit{pseudo label mixing} improves the performance in all scenarios instead in \textit{phoneme level mix} with ratio 25\%. This might be due to the trade-off between utilizing more data and the distribution mismatch since the variance adaptor and decoder are tuned on a mixed sequence from different encoders' outputs. When the ratio is low, the model relies too much on $c_{repr}$ during fine-tuning. Finally, although applying \textit{mix pretraining} itself does not incur too much improvement, it is highly effective when combined with \textit{pseudo label mixing}.

For the Mean of Opinion Score test, we fixed the pseudo label ratio to 75\% and randomly sampled 20 different sentences from every experiment. The test is conducted on Prolific platform, where each sentence is rated by at least 5 individuals, and over 30 native speakers are invited for both languages. The ground truth audios are also evaluated for reference. Ground truth audios are first transformed into Mel-spectrograms and then resynthesized back to audios to remove the influence of the vocoder. The results are summarized in Table~\ref{tab:main-mos-exp}.

\subsection {Varying Adaptation Data}

To verify the data efficiency of the proposed method, we experiment with different extremely low-resource data settings. We set the amount of $D_{target}$ to 4, 16, and 64 utterances, and the amount of unpaired speech data from 0, 15, 60 to 240 minutes. We compare the following 3 methods. \textbf{(a)} \textbf{Baseline}: No mix pretraining, sentence-level filter, pseudo label ratio 100\%. \textbf{(b)} \textbf{Proposed}: Mix pretraining, phoneme-level mix, pseudo label ratio 75\%. \textbf{(c)} \textbf{Proposed*}: \textbf{Proposed} extended with embedding initialization trick in Section~\ref{section:FSCL}.

\begin{figure}[htb]
\begin{minipage}[b]{1.0\linewidth}
  \centering
  \centerline{\includegraphics[width=1.0\linewidth]{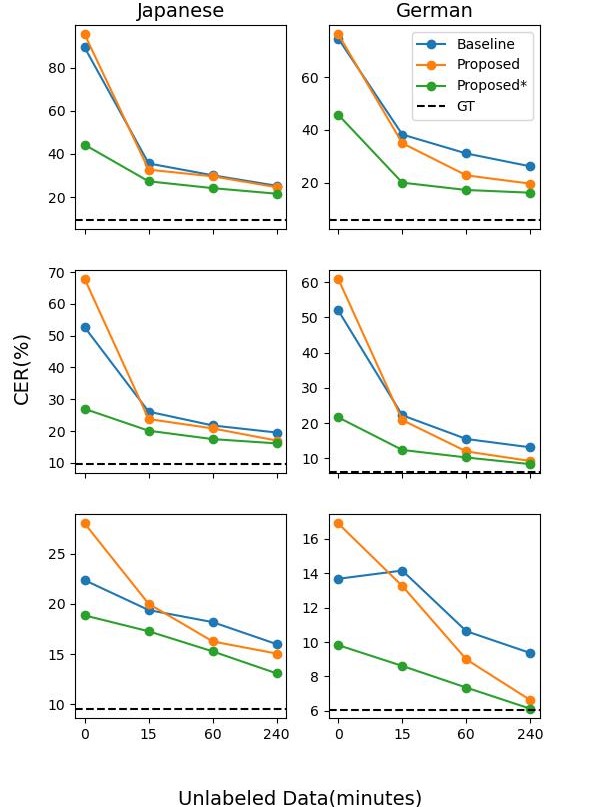}}
\end{minipage}
\caption{CER[\%] under different data settings. Different rows represent different amounts of $D_{target}$. From the first row to the last row are 4-shot, 16-shot, and 64-shot.}
\label{fig:data1}
\end{figure}

CER is averaged over 5 runs, and ground truth's CER is also provided for comparison. The results are shown in Figure~\ref{fig:data1}. In the absence of unpaired speech data, only \textbf{Proposed*} is barely able to adapt to the new language. Since only the embedding initialization is applied in this case, the results can be seen as results of \cite{huang2022fewshot}. Compared to that, \textbf{Proposed*} further exploits unpaired speech data and significantly improves by utilizing just 15 minutes of unpaired speech data. Remarkably, with only 64 utterances of paired data ($\sim$10 minutes) and 4 hours of unpaired speech data, \textbf{Proposed*} achieves comparable CER to the ground truth in German's case. In different data settings, \textbf{Proposed} consistently outperforms \textbf{Baseline}, while \textbf{Proposed*} surpasses both, especially in low-resource scenarios where good initialization plays a critical role. The results demonstrate that \textbf{Proposed*} is a highly data-efficient adaptation method.

\subsection{Further Analyses}

\subsubsection{Mixing Function}

There are various ways to construct the mixing function. An alternative way is to make the decision with probability instead of deterministically according to confidence. Denote a function $s(x, y;\alpha)$ such that output is $x$ with probability $\alpha$, $y$ with probability $1-\alpha$. Consider two mixing functions:

\noindent {\bfseries Soft Mix.}\quad For every phoneme, this method decides whether to use $c_{phn}$ or $c_{repr}$ with high probability based on if its confidence score has surpassed a predefined threshold. The mixing function is
\begin{equation}
c_{mix}^i = f(c_{repr}^i, c_{phn}^i) = \begin{cases} 
s(c_{phn}^i, c_{repr}^i;\alpha),&\text{if }s^i\geq \lambda \\
s(c_{phn}^i, c_{repr}^i;1 - \alpha),&\text{otherwise},
\end{cases}
\end{equation}
where $\lambda, \alpha$ are hyperparameters.

\noindent {\bfseries Sampling.}\quad For every phoneme, this method selects $c_{phn}$ with probability equals to its confidence score. The mixing function is
\begin{equation}
c_{mix}^i = f(c_{repr}^i, c_{phn}^i) = s(c_{phn}^i, c_{repr}^i;s^i).
\end{equation}

Table~\ref{tab:extra-exp1} compares proposed \textit{phoneme-level mix} with the two methods. We follow the experimental setup in Section~\ref{section:eval1}. \textit{Phoneme-level mix} is renamed as \textbf{Hard Mix} for clarification since all three methods operate on the phoneme level. \textit{Mix pretraining} is applied. Note that \textbf{Sampling} has its own pseudo label ratio by definition, so we are unable to assign a pseudo label ratio. 

\begin{table}[t]
  \caption{CERs[\%] of different mixing functions. \textbf{Sampling} has its own pseudo label ratio by definition, so we are unable to assign a pseudo label ratio. }
  \centering
  \begin{tabular}{cc|ccc}
    \toprule
    \multirow{2}{*}{\textbf{Lang.}} & \multirow{2}{*}{\textbf{FT}} & \multicolumn{3}{c}{\textbf{Pseudo Label Ratio}}\\
    & & 25\% & 50\% & 75\% \\
    \midrule
    \multirow{3}{*}{\textit{jp}} & Hard Mix & 21.66 & 17.47 & 17.23 \\
    & Soft Mix ($\alpha=0.9$) & 23.82 & 19.66 & 18.05 \\
    \cmidrule{2-5}
    & Sampling & \multicolumn{3}{c}{17.25} \\
    \midrule
    \multirow{3}{*}{\textit{de}} & Hard Mix & 15.23 & 11.31 & 9.44 \\
    & Soft Mix ($\alpha=0.9$) & 15.18 & 12.55 & 10.65 \\
    \cmidrule{2-5}
    & Sampling & \multicolumn{3}{c}{11.47} \\
    \bottomrule
  \end{tabular}
  \label{tab:extra-exp1}
\end{table}

Results show that \textbf{Hard Mix} with 75\% pseudo label ratio achieves the best performance in both languages. However, in practice, we do not know the perfect pseudo label ratio in advance, therefore \textbf{Sampling} becomes a practical choice since it is good enough and avoids threshold selection.

\subsubsection{Pseudo Label Quality}

We investigate how different ASR performance, or the quality of pseudo labels, affects the proposed method. For the pseudo label generation, we tune the linear head with 4, 16, and 64 utterances, resulting in different pseudo labels. Following Section~\ref{section:eval1}, we tune the TTS with 16 utterances of labeled data and the pseudo labels. Table~\ref{tab:extra-exp2} compares the best and the worst adaptation strategy combination from Section~\ref{section:strat}, where pseudo label ratio is fixed to 75\%. Experiment results show that the best combination, \textit{mix pretraining} with \textit{phoneme level mix}, stays effective under all settings.

\begin{table}[t]
  \caption{CERs[\%] of TTS tuned with different pseudo labels.}
  \centering
  \begin{tabular}{cc|ccc}
    \toprule
    \multirow{2}{*}{\textbf{Lang.}} & \multirow{2}{*}{\textbf{Stragegy}} & \multicolumn{3}{c}{\textbf{ASR Data Usage}}\\
    & & 4-shot & 16-shot & 64-shot \\
    \midrule
    \multirow{2}{*}{\textit{jp}} & w/ MP + PM & 19.23 & 17.23 & 13.14 \\
    & w/o MP + SF & 30.24 & 28.59 & 15.54 \\
    \midrule
    \multirow{2}{*}{\textit{de}} & w/ MP + PM & 18.80 & 9.44 & 7.56 \\
    & w/o MP + SF & 25.82 & 13.4 & 10.73 \\
    \bottomrule
  \end{tabular}
  \label{tab:extra-exp2}
\end{table}

\section{Conclusion}

In this paper, we propose a highly data-efficient transfer learning framework for language adaptation in TTS. Our method incorporates self-supervised features during both the pretraining and fine-tuning stages, allowing us to leverage more information from the data compared to the naive baseline. 
Additionally, by extending our method with an embedding initialization trick, the model can adapt to an unseen language with only 4 utterances of labeled data and 15 minutes of unlabeled data. 
We demonstrate the effectiveness of our approach, showcasing significant improvements in various extremely low-resource settings.

\section{Acknowledgements}
We thank to National Center for High-performance Computing (NCHC) of National Applied Research
Laboratories (NARLabs) in Taiwan for providing computational and storage resources.

\bibliographystyle{IEEEbib}
\bibliography{strings,refs}

\end{document}